\pdfoutput=1

\documentclass[11pt]{article}

\usepackage[]{naacl2021}

\usepackage{times}
\usepackage{latexsym}

\usepackage[T1]{fontenc}
\usepackage{booktabs}

\usepackage[utf8]{inputenc}

\usepackage{microtype}

\usepackage{graphicx}
\usepackage{todonotes}
\usepackage{stfloats}
\usepackage[flushleft]{threeparttable} 
\usepackage{tcolorbox}

\usepackage{hyperref}
\usepackage{float}

\newcommand\blfootnote[1]{%
  \begingroup
  \renewcommand\thefootnote{}\footnote{#1}%
  \addtocounter{footnote}{-1}%
  \endgroup
}

\title{How Many Data Points is a Prompt Worth?}

\author{Teven Le Scao \\
  Hugging Face \\
  \texttt{teven@huggingface.co} \\\And
  Alexander M. Rush \\
  Hugging Face \\
  \texttt{sasha@huggingface.co} \\}
\usepackage{lipsum}
\begin{document}
\maketitle
\begin{abstract}
When fine-tuning pretrained models for classification, researchers either use a generic model \textit{head} or a task-specific \textit{prompt} for prediction. Proponents of prompting have argued that prompts provide a method for injecting task-specific guidance, which is beneficial
in low-data regimes.
We aim to quantify this benefit through rigorous testing of prompts 
in a fair setting: comparing prompted and head-based fine-tuning in equal conditions across many tasks and data sizes. 
By controlling for many sources of advantage, we find that prompting does indeed provide a benefit, and that this benefit can be quantified 
per task. Results show that prompting is often worth 100s of data points on average across classification tasks.

\end{abstract}

\section{Introduction}







The main paradigm for adapting pretrained models for classification~\cite{GPT, UniLM, BERT} is fine-tuning via an explicit classifier head. However, an alternative approach has arisen: adapting the pretrained language model directly as a predictor through autoregressive text generation~\cite{GPT2} or completion of a cloze task~\cite{ASimpleMethod}. This method is notably used in T5 fine-tuning~\cite{T5} leading to state-of-the-art results on the SuperGLUE benchmark~\cite{SuperGLUE}.

\blfootnote{Code available at \url{https://github.com/TevenLeScao/pet}}


One argument made for classification by direct language generation is that it allows us to pick custom \textit{prompts} for each task~\cite{Decathlon}. 
While this approach can be used for zero-shot classification~\cite{Zeroshot} or priming~\cite{GPT3}, it can also be used in fine-tuning to provide extra task information to the classifier, especially in the low-data regime~\cite{PET, PET2}.

If this argument is indeed true, it is natural to ask how it impacts the sample efficiency of the model, or more directly, \textit{how many data points is a prompt worth?} As with many low-data and pretraining-based problems, this 
question is complicated by the fine-tuning setup, training procedure, and prompts themselves. We attempt to isolate these variables through diverse prompts, multiple runs, and best practices in low-training data fine-tuning. We introduce a metric, the \textit{average data advantage}, for quantifying the impact of a prompt in practice.


Our experiments find that the impact of task-targeted prompting can nicely be quantified in terms of direct training data, and that it varies over the nature of different tasks. On MNLI~\cite{MNLI}, we find that using a prompt contributes approximately 3500 data points. On SuperGLUE, it adds approximately 280 data points on RTE~\cite{RTE} and up to 750 on BoolQ~\cite{BoolQ}. In low- to medium-data settings, this advantage can be a real contribution to training a model. 


\section{Related Work}


Prompting has been used both for zero-shot and fine-tuning based methods. Zero-shot  approaches attempt to answer a task with a prompt without fine-tuning through generation~\cite{GPT2}. GPT3~\cite{GPT3} extends this approach to a supervised priming method by taking in training data as priming at inference time, so it can attend to them while answering. T5~\cite{T5} and other sequence-to-sequence pretrained models use standard word-based fine-tuning with a marker prompt to answer classification tasks with strong empirical success. Our setting differs in that we are interested in using task-based prompts and fine-tuning, in-between the T5 and GPT2 setting. 

Our setting most closely resembles PET~\cite{PET, PET2}, which claims that task-specific prompting helps transfer learning, especially in the low-data regime. However, in order to reach the best possible results on SuperGLUE, PET introduces several other extensions: semi-supervision via additional pseudo-labeled data, ensembling models trained with several different prompts, and finally distilling the ensemble into a linear classifier rather than a language model. Our aim is to isolate the specific contributions of prompting within supervised fine-tuning. 

Finally, recent papers have experimented with discovering prompts through automated processes tailored to the language model~\cite{HowCanWeKnow, AutomaticVerbalizer}. We limit ourselves to human-written prompts, as we are interested into 
whether prompting itself specifically adds information to the supervised task. It is an interesting question as to whether automatic prompts can have this same impact (relative to the training data they require). 

\section{Comparison: Heads vs Prompts}



Consider two transfer learning settings for text classification: \textit{head-based}, where a generic head 
layer takes in pretrained representations to predict an output class; \textit{prompt-based}, where a 
task-specific pattern string is designed to coax the model into producing a textual output corresponding to 
a given class. Both can be utilized for fine-tuning with supervised training data, but prompts further allow
the user to customize patterns to help the model. 

For the \textit{prompt} model we follow the notation from PET~\cite{PET} and decompose a prompt into a \textit{pattern} and a \textit{verbalizer}. The \textit{pattern} turns the input text into a cloze task, i.e. a sequence with a masked token or tokens that need to be filled. Let us use as example an excerpt from SuperGLUE task BoolQ~\cite{BoolQ}, in which the model must answer yes-or-no binary questions. In order to let a language model answer the question in \textit{italics}, our pattern is in \textbf{bold}~\cite{PET2}:

\begin{quote}
\small
 "Posthumous marriage -- Posthumous marriage (or necrogamy) is a marriage in which one of the participating members is deceased. It is legal in France and similar forms are practiced in Sudan and China. Since World War I, France has had hundreds of requests each year, of which many have been accepted.
\textbf{Based on the previous passage, \textit{can u marry a dead person in france ?} <MASK>}"
\end{quote}



The masked word prediction is mapped to a \textit{verbalizer} which produces a class. (here "Yes": True. "No": False\footnote{The correct answer here is, of course, \textit{yes}. Originated in 1803 as Napoleon rose to power, this practice was mainly to the benefit of war widows.}). 
Several \textit{pattern-verbalizer pairs} (\textit{PVPs}) could be used for a single task, differing either through the pattern, the verbalizer, or both. Fine-tuning is done by training the model to produce the correct verbalization. The loss is the cross-entropy loss between the correct answer and the distribution of probabilities amongst the tokens in the verbalizer. We re-use pattern choices from~\citet{PET2}; examples are available in Appendix~\ref{prompts}.




\section{Experimental Setting}


We run all experiments with the same pretrained checkpoint, \textit{roberta-large} (355M parameters) from RoBERTa~\cite{Roberta}, which we load from the \textit{transformers}~\cite{Transformers} library.\footnote{After experimenting with RoBERTa, AlBERT~\cite{Albert} and BERT~\cite{BERT}, we found \textit{roberta-large} to have the most consistent performance.}
In line with previous observations~\cite{Feather,Finetuning,Mixout}, head-based fine-tuning performance varies considerably. We follow recommendations of~\citet{Stability} and~\citet{Revisiting} to train at a low learning rate ($10^{-5}$) for a large number of steps (always at least $250$, possibly for over 100 epochs).





\begin{figure*}[h!]
\centering
\hspace*{-1cm}\includegraphics[width=1.1\textwidth]{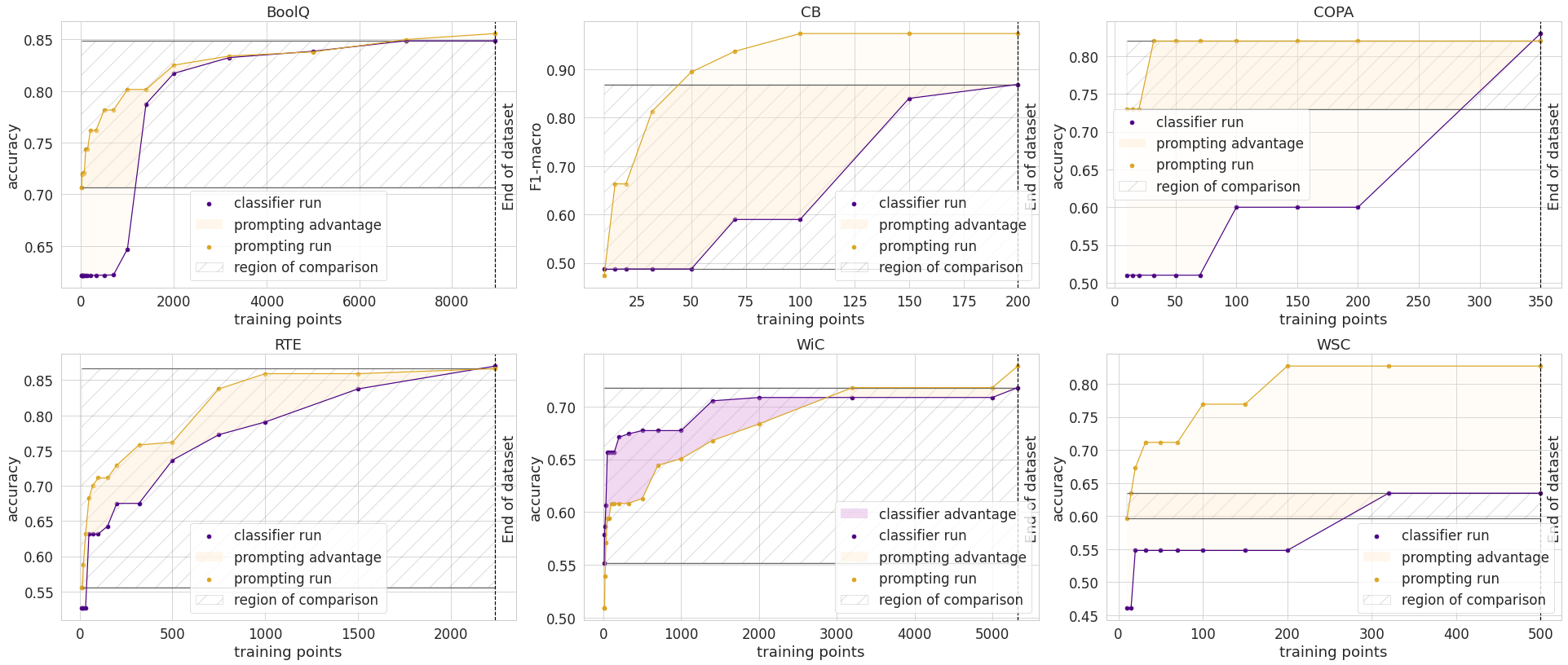}
\caption{Prompting vs head (classifier) performance across data scales, up to the full dataset, for six SuperGLUE tasks. Compares the best prompt and head performance at each level of training data across 4 runs. Highlighted region shows the accuracy difference of the models. Cross-hatch region highlights the lowest- and highest- accuracy matched region in the curves. The highlighted area in this region is used to estimate the data advantage. }
\label{main_figure}
\end{figure*}

We perform our evaluation on SuperGLUE and MNLI~\cite{MNLI}. These datasets comprise a variety of tasks, all in English, including entailment (MNLI, RTE~\cite{RTE}, CB~\cite{CB}), multiple choice question answering (BoolQ~\cite{BoolQ}, MultiRC~\cite{MultiRC}), and common-sense reasoning (WSC~\cite{WSC}, COPA~\cite{COPA}, WiC~\cite{WiC}). We do not include ReCoRD~\cite{ReCORD} in our comparisons as there is no head model to compare with, since it is already a cloze task. Data sizes range from $250$ data points for CB to $392,702$ for MNLI. As test data is not publicly available for SuperGLUE tasks, we set aside part of training (from $50$ for CB, COPA and MultiRC to $500$ for BoolQ) to use for development, and evaluate on their original validation sets. For MNLI, we use the available matched validation and test sets. 

We compare models across a scale of available data, starting with $10$ data points and increasing exponentially (as high-data performance tends to saturate) to the full dataset. For example, for MultiRC, which has 969 data points initially, we start by reserving 50 data points for development. This leaves us with 919 training points, and we train models with 10, 15, 20, 32, 50, 70, 100, 150, 200, 320, 500, 750, and 919 training points. We run every experiment 4 times in order to reduce variance, for a total of 1892 training runs across all tasks. At every point, we report the best performance that has been achieved at that amount of data or lower. Full graphs are presented in Appendix~\ref{reduction}.


\section{Results}

Figure~\ref{main_figure} shows the main results comparing head- and prompt-based fine-tuning with the best-performing pattern on that task. 
Prompting enjoys a substantial advantage on every task, except for WiC as is reported in previous results~\cite{PET2}.
Both approaches improve with more training data, but prompting remains better by a varying amount. Many tasks in SuperGLUE have relatively few data points, but we also see an advantage in large datasets like BoolQ and MNLI.

To quantify how many data points the prompt is worth, we first isolate the $y$-axis band of the lowest- and highest- accuracy where the two curves match in accuracy.\footnote{We assume asymptotically the two curves would match, but are limited by data.} The horizontal line at these points represents the advantage of prompting. We then take the integral in this region, i.e. area between the linearly-interpolated curves\footnote{In areas where the head model is better, if any, get subtracted from the total.}, divided by the height of the band. The area has the dimension of a quantity of data points times the metric unit, so dividing by the performance range yields a \# of data points advantage. 
As low data training is sensitive to noise, in addition to following best training practices we 
run several different experiments for each $x$-point. We use a bootstrapping approach to estimate confidence over these runs. Specifically, we hold out one of the 4 head runs and 4 prompt runs (16 combinations total), and compute the standard deviation of those outcomes.

We report these quantities for every task in Table~\ref{main_table} as \textit{Average advantage}. For almost all the tasks, we see that prompting gives a substantial advantage in terms of data efficiency, adding the equivalent of hundreds of data points on average.


\begin{table*}

\hspace*{-0.7cm}\begin{tabular}{@{}l rrrrrrrr@{}}
\toprule 
& \multicolumn{8}{c}{Average Advantage (\# Training Points)} \\
& MNLI & BoolQ & CB & COPA & MultiRC* & RTE & WiC & WSC\\\midrule \multicolumn{1}{l}{\textit{P vs H}} & $3506\pm536$ & $752\pm46$ & $90\pm2$ & $288\pm242$ & $384\pm378$  & $282\pm34$ & $-424\pm74$ & $281\pm137$ \\
\hline
\multicolumn{1}{l}{\textit{P vs N}} & $150\pm252$ & $299\pm81$ & $78\pm2$ & -& $74\pm56\phantom{0}$ & $404\pm68$ & $-354\pm166$ & -\\
\multicolumn{1}{l}{\textit{N vs H}} & $3355\pm612$ & $453\pm90$ & $12\pm1$ & -& $309\pm320$ & $-122\pm62$ & $-70\pm160$ & -\\
\bottomrule
\end{tabular}

\caption{Average prompting advantage in number of data points for MNLI \& SuperGLUE tasks. \textit{P} denotes the prompt model, \textit{H} the head model. On average across performance levels, an MNLI prompt model yields the results of an MNLI head model trained with 3500 additional data points. Confidence levels are based on a multiple random runs (see text). \textit{N} indicates a null-verbalizer prompting task that replaces the verbalizer with a non-sensical mapping.  *The comparison band of MultiRC is too small as the head baseline fails to learn beyond majority class; we use the full region for a lower-bound result.}
\label{main_table}
\end{table*}




\section{Analysis}

\paragraph{Impact of Pattern vs Verbalizer}

The intuition of prompts is that they introduce a task description in natural language,
even with few training points. 
To better understand the zero-shot versus adaptive nature of prompts,
we consider a \textit{null verbalizer}, a control with a verbalizer that cannot yield semantic information without training. For every task that requires filling in one word (which excludes the more free-form COPA and WSC), we replace the verbalizers, for example, "yes", "no", "maybe", "right" or "wrong",  with random first names.

\begin{figure}
\centering
\includegraphics[width=0.8\columnwidth]{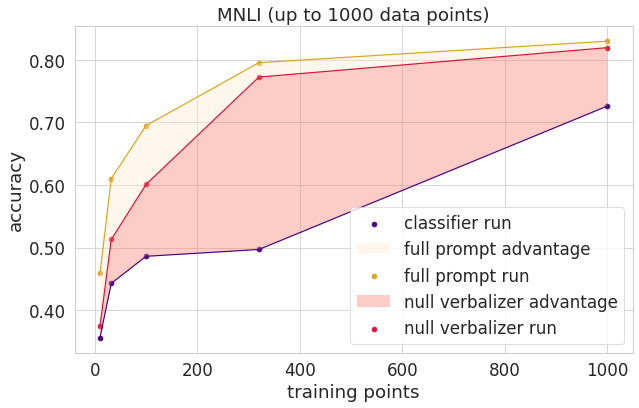}
\caption{Comparison of full prompt and null verbalizer advantage on MNLI at lower data scales.}
\label{neutral_run_figure}
\end{figure}

Table~\ref{main_table} shows the advantage of the standard prompt over the null verbalizer to estimate this control.
We see that for small data tasks such as CB, the null verbalizer removes much of the benefits of prompting. However, with more training data, the model seems to adapt the verbalizer while still gaining the inductive bias benefits of the pattern. Figure~\ref{neutral_run_figure} showcases this dynamic on MNLI. This result further shows that prompting yields data efficiency even if it is not directly analogous to the generation process of training. 


\paragraph{Impact of Different Prompts}

If the prompt acts as a description of the task, one would expect different valid descriptions to vary in their benefits. In order to compare the different prompts we used on each task, we chart the median performance for each of them under different runs. In nearly every experiment, we find that the confidence intervals of those curves largely overlap, implying that prompt choice is not a dominant hyperparameter, i.e. the variance across random seeds usually outweighs the possible benefits of prompt choice. One exception is the low-data regime of BoolQ, where one of the prompts enjoys a significant few-shot advantage over the others. We plot this curve for MultiRC in Figure~\ref{median_comparison} and the rest in Appendix~\ref{all_results}.

\begin{figure}
\centering
\includegraphics[width=0.8\columnwidth]{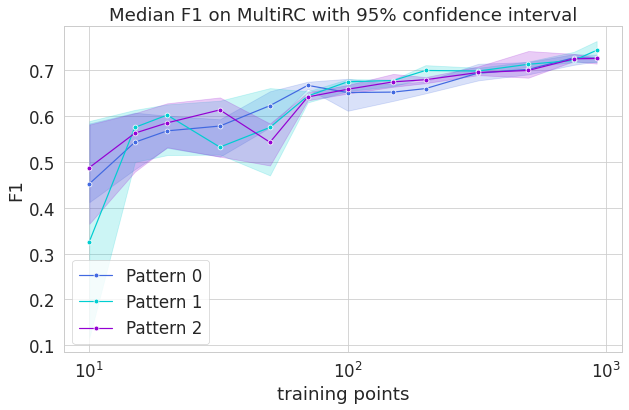}
\caption{Median performance on MultiRC across runs for three prompts. Differences are inconsistent and eclipsed by the variance within one prompt's runs.}
\label{median_comparison}
\end{figure}

\paragraph{Metric sensitivity}

We treat each metric linearly in calculating advantage; alternatively, we could re-parameterize the $y$ axis for each task. This choice does not have a consistent effect for or against prompting. For example, emphasizing gains close to convergence increases prompting advantage on CB and MNLI but decreases it on COPA or BoolQ. 


\section{Conclusion}

We investigate prompting through a systematic study of its data advantage. Across tasks, prompting consistently yields a varying improvement throughout the training process. Analysis shows that prompting is mostly robust to pattern choice, and can even learn without an informative verbalizer. On large datasets, prompting is similarly helpful in terms of data points, although they are less beneficial in performance. In future work, we hope to study the mechanism and training dynamics of the prompting benefits.

\section{Impact statement}

Significant compute resources were used to run this paper's experiments. A single experiment (defined as one model run, at one data level, on one task) was quite light-weight, taking usually a little under an hour on a single Nvidia V100. However, as we computed a little under two thousand runs, this adds up to about 1800 GPU hours, to which one must add around 400 GPU hours of prototyping and hyper-parameter searching. Those 2200 GPU hours would usually have necessitated the release of about 400kg of CO2, about 40\% of a transatlantic flight for a single passenger, in the country where we ran the experiments, although we used a carbon-neutral cloud compute provider.

The main benefit of prompting, rather than compute efficiency, is data efficiency. Although we ran all of our experiments on English, we hope that this property will be especially helpful in low-resource language applications. In a sense, a practitioner could then remedy the lack of task-specific data in their language by introducing information through a prompt. However, this comes with the inherent risk of introducing human biases into the model. Prompt completion also suffers from biases already present within the language model~\cite{Babysitter}. This could cause a prompted model to repeat those biases in classification, especially in the few-shot setting where prompting mostly relies on the pretrained model.

\section{Acknowledgments}

We thank Steven Cao and Joe Davison for the discussions about prompting that initially spurred this paper. We further thank Timo Schick for making the code for PET available and for discussions about performance replication. We lastly thank Canwen Xu, Yacine Jernite, Victor Sanh, Dimitri Lozeve and Antoine Ogier for their help with the figures and writing of this draft.


\bibliography{custom}
\bibliographystyle{acl_natbib}

\newpage
\appendix

\section{Choice of prompts}
\label{prompts}

We use a subset of prompts from~\cite{PET2}.
\begin{itemize}
    \item For entailment tasks (\textbf{CB}, \textbf{MNLI}, \textbf{RTE}) given a premise $p$ and hypothesis $h$: 
    \begin{tcolorbox}[width=0.9\columnwidth, colback=white!95!black]\small h? | <MASK>, p\end{tcolorbox} \begin{tcolorbox}[width=0.9\columnwidth, colback=white!95!black]\small "h"? | <MASK>. "p"\end{tcolorbox}
    with "yes" as a verbalizer for entailment, "no" for contradiction, "maybe" for neutrality.

    \item For \textbf{BoolQ}, given a passage $p$ and question $q$:
    \begin{tcolorbox}[width=0.9\columnwidth, colback=white!95!black]\small $p$. Question: $q$? Answer: <MASK>.\end{tcolorbox}
    \begin{tcolorbox}[width=0.9\columnwidth, colback=white!95!black]\small $p$. Based on the previous passage, $q$? <MASK>.\end{tcolorbox}
    \begin{tcolorbox}[width=0.9\columnwidth, colback=white!95!black]\small Based on the following passage, $q$? <MASK>. $p$\end{tcolorbox}
    with "yes" or "no" as verbalizers for True and False.

    \item For \textbf{COPA}, given an effect $e$ and possible causes $c_1$ and $c_2$:
    \begin{tcolorbox}[width=0.9\columnwidth, colback=white!95!black]\small “$c_1$” or “$c_2$”? $e$, so <MASK> .\end{tcolorbox}
    \begin{tcolorbox}[width=0.9\columnwidth, colback=white!95!black]\small $c_1$ or $c_2$? $e$, so <MASK>.\end{tcolorbox}
    and a cause $c$ and possible effects $e_1$ and $e_2$:
    \begin{tcolorbox}[width=0.9\columnwidth, colback=white!95!black]\small “$e_1$” or “$e_2$”? <MASK>, because $c$.\end{tcolorbox}
    \begin{tcolorbox}[width=0.9\columnwidth, colback=white!95!black]\small $e_1$ or $e_2$? <MASK>, because $c$.\end{tcolorbox}
    and the verbalizer is the identity function.
    
    \item For \textbf{MultiRC}, given a passage $p$, question $q$ and answer $a$, we estimate whether $a$ is a proper answer with:
    \begin{tcolorbox}[width=0.9\columnwidth, colback=white!95!black]\small $p$. Question: $q$? Is it $a$? <MASK>.\end{tcolorbox}
    \begin{tcolorbox}[width=0.9\columnwidth, colback=white!95!black]\small $p$. Based on the previous passage, $q$? Is the correct answer $a$? <MASK>.\end{tcolorbox}
    \begin{tcolorbox}[width=0.9\columnwidth, colback=white!95!black]\small $p$. Question: $q$? Is the correct answer $a$? <MASK>.\end{tcolorbox}
    
    \item For \textbf{WiC}, given two sentences $s_1$ and $s_2$ and a word $w$, we classify whether $w$ was used in the same sense.
    \begin{tcolorbox}[width=0.9\columnwidth, colback=white!95!black]\small "$s_1$” / “$s_2$”. Similar sense of “$w$”? <MASK>.\end{tcolorbox}
    \begin{tcolorbox}[width=0.9\columnwidth, colback=white!95!black]\small $s_1$ $s_2$ Does $w$ have the same meaning in both sentences? <MASK>.\end{tcolorbox}
    With "yes" and "no" as verbalizers.
    
    \item For \textbf{WSC}, given a sentence $s$ with a marked pronoun *$p$* and noun $n$:
    \begin{tcolorbox}[width=0.9\columnwidth, colback=white!95!black]\small $s$ The pronoun ‘*$p$*’ refers to <MASK>.\end{tcolorbox}
    \begin{tcolorbox}[width=0.9\columnwidth, colback=white!95!black]\small $s$ In the previous sentence, the pronoun ‘*$p$*’refers to <MASK>.\end{tcolorbox}
    \begin{tcolorbox}[width=0.9\columnwidth, colback=white!95!black]\small $s$ In the passage above, what does the pronoun‘*$p$*’ refer to? Answer: <MASK>.\end{tcolorbox}
    With the identity function as a verbalizer.
    
\end{itemize}

\section{Influence of the reporting method over runs}
\label{reduction}

We chose to report the \textbf{accumulated maximum} performance across runs for every model. This means that if the maximum performance over random seeds is smaller than a maximum previously attained with less data points, we use the previous value. This appendix presents results with the \textbf{maximum} and \textbf{mean} at every point to condense several runs instead. Using either maximum is equivalent; using the mean, however, can make results vary significantly, as the distribution of outcomes is heavily left-skewed, or even bimodal, with poor-performance outliers.

\begin{table*}

\hspace*{-0.7cm}\begin{tabular}{@{}l rrrrrrrr@{}}
\toprule 
& \multicolumn{8}{c}{Average Advantage (accumulated maximum reporting)} \\
& MNLI & BoolQ & CB & COPA & MultiRC* & RTE & WiC & WSC\\\midrule \multicolumn{1}{l}{\textit{P vs H}} & $3506\pm536$ & $752\pm46$ & $90\pm2$ & $288\pm242$ & $384\pm378$  & $282\pm34$ & $-424\pm74$ & $281\pm137$ \\
\hline
\multicolumn{1}{l}{\textit{P vs N}} & $150\pm252$ & $299\pm81$ & $78\pm2$ & -& $74\pm56\phantom{0}$ & $404\pm68$ & $-354\pm166$ & -\\
\multicolumn{1}{l}{\textit{N vs H}} & $3355\pm612$ & $453\pm90$ & $12\pm1$ & -& $309\pm320$ & $-122\pm62$ & $-70\pm160$ & -\\
\bottomrule
\end{tabular}

\caption{Average prompting advantage in number of data points for MNLI \& SuperGLUE tasks with accmax reporting. \textit{P} denotes the prompt model, \textit{H} the head model. On average across performance levels, an MNLI prompt model yields the results of an MNLI head model trained with 3500 additional data points. Confidence levels are based on a multiple random runs (see text). \textit{N} indicates a null-verbalizer prompting task that replaces the verbalizer with a non-sensical mapping.  *The comparison band of MultiRC is too small as the head baseline fails to learn beyond majority class; we use the full region for a lower-bound result.}
\label{appendix_accmax}
\end{table*}

\begin{table*}

\hspace*{-0.7cm}\begin{tabular}{@{}l rrrrrrrr@{}}
\toprule 
& \multicolumn{8}{c}{Average Advantage (maximum reporting)} \\
& MNLI & BoolQ & CB & COPA & MultiRC* & RTE & WiC & WSC\\\midrule \multicolumn{1}{l}{\textit{P vs H}} & $3506\pm536$ & $737\pm53$ & $86\pm1$ & $226\pm189$ & $448\pm314$  & $218\pm24$ & $-133\pm65$ & $297\pm448$ \\
\hline
\multicolumn{1}{l}{\textit{P vs N}} & $150\pm252$ & $249\pm80$ & $75\pm3$ & -& $77\pm64\phantom{0}$ & $331\pm64$ & $-341\pm187$ & -\\
\multicolumn{1}{l}{\textit{N vs H}} & $3355\pm612$ & $488\pm90$ & $12\pm2$ & -& $371\pm305$ & $-113\pm61$ & $209\pm176$ & -\\
\bottomrule
\end{tabular}

\caption{Average prompting advantage in number of data points for MNLI \& SuperGLUE tasks with maximum reporting.}
\label{appendix_max}
\end{table*}

\begin{table*}

\hspace*{-0.7cm}\begin{tabular}{@{}l rrrrrrrr@{}}
\toprule 
& \multicolumn{8}{c}{Average Advantage (mean reporting)} \\
& MNLI & BoolQ & CB & COPA & MultiRC* & RTE & WiC & WSC\\\midrule \multicolumn{1}{l}{\textit{P vs H}} & $4526\pm266$ & $1139\pm285$ & $81\pm2$ & $292\pm212$ & $399\pm77$  & $72\pm71$ & $447\pm190$ & $-490\pm402$ \\
\hline
\multicolumn{1}{l}{\textit{P vs N}} & $185\pm307$ & $514\pm287$ & $80\pm2$ & -& $58\pm2$ & $621\pm215$ & $10\pm104$ & -\\
\multicolumn{1}{l}{\textit{N vs H}} & $4341\pm347$ & $625\pm402$ & $1\pm2$ & -& $342\pm78$ & $-549\pm203$ & $437\pm189$ & -\\
\bottomrule
\end{tabular}

\caption{Average prompting advantage in number of data points for MNLI \& SuperGLUE tasks with mean reporting.}
\label{appendix_mean}
\end{table*}

\newpage

\section{Curves on all tasks}
\label{all_results}

\begin{figure*}[b]
\centering
\hspace*{-1cm}\includegraphics[width=1\textwidth]{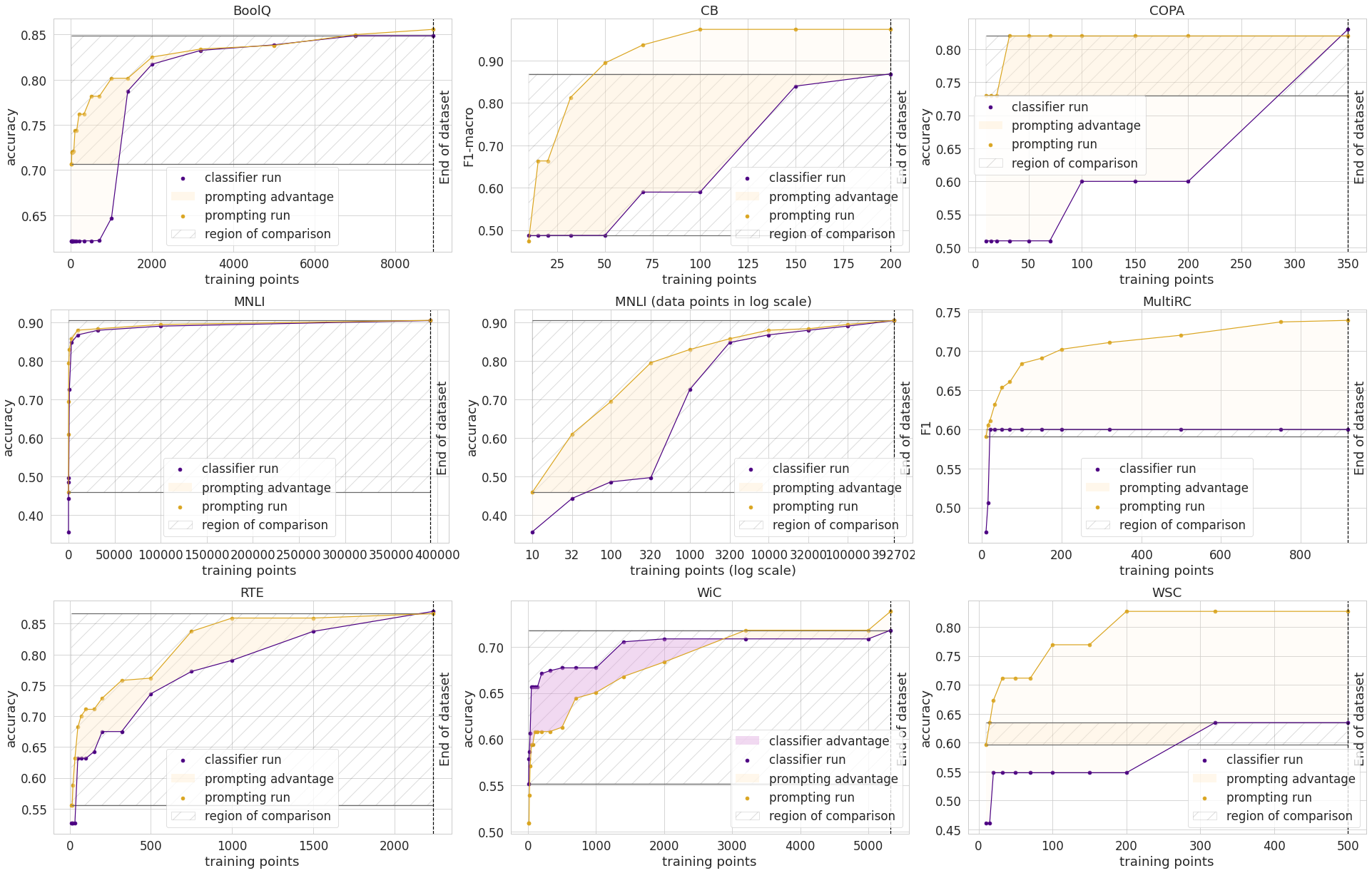}
\caption{Prompting vs head (classifier) performance across data scales, up to the full dataset, for seven SuperGLUE tasks \& MNLI. Compares the best prompt and head performance at each level of training data across 4 runs. Highlighted region shows the accuracy difference of the models. Cross-hatch region highlights the lowest- and highest- accuracy matched region in the curves. The highlighted area in this region is used to estimate the data advantage from prompting.}
\label{appendix_master}
\end{figure*}

\restoregeometry

\begin{figure*}
\centering
\hspace*{-1cm}\includegraphics[width=1\textwidth]{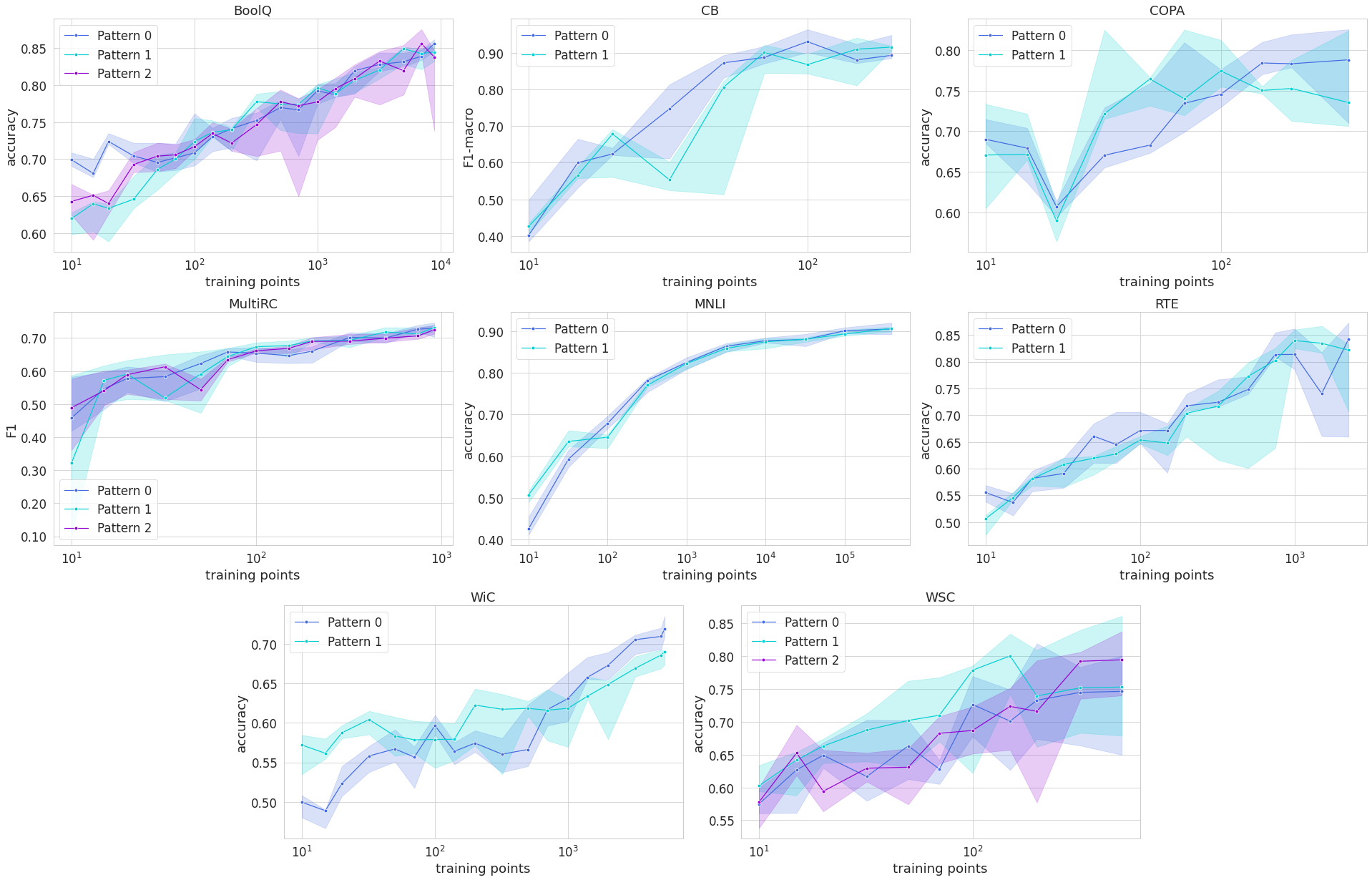}
\caption{Median performance across runs for each prompt on every task.}
\label{appendix_median}
\end{figure*}

\end{document}